# Correlation Is Not Enough: Embedding Human Metadata for Individual Causal Discovery

*Cross-domain failure in biomedical embeddings, ontology-graph fine-tuning, and AMX-native inference on Intel Xeon*

**Suraj Biswas**[1,2] **Saurav Gupta**[1,2] **Pritam Mukherjee**[1,2]
[1] *Assessli Research* [2] *Dots-In Research, Bengaluru, India*
ceo@dotsin.ai · cto@dotsin.ai · pritam.mukherjee@assessli.com
ORCID · Suraj Biswas 0009-0008-1727-8179 · Pritam Mukherjee 0009-0007-9018-4083

## Abstract

Ask a pretrained biomedical language model whether "cortisol 28 µg/dL" and "stock-market volatility" are related, and it returns a cosine similarity of 0.83 on a scale where 1.0 means identical. The two share no mechanism. This is not a corner case. Every off-the-shelf biomedical encoder we tested — BioBERT, PubMedBERT, BioM-ELECTRA — scores unrelated pairs from different domains between 0.76 and 0.92 when the answer should be near zero. On a six-part diagnostic, accuracy on cross-domain discrimination is exactly 0%.

Most retrieval systems survive this, because a language model downstream filters the noise. A Large Behavioural Model (LBM) — a foundation model whose subject is a person, not a sentence — does not. An LBM reasons over a graph of a user's life and reads embedding proximity as evidence that two events are causally linked. If the geometry says cortisol and the stock market belong together, the model writes a false causal edge, and everything downstream inherits the mistake. For this system, embedding geometry is not a tuning knob. It is correctness.

We report the fix and its price. A first contrastive pass over 72,034 pairs raises PubMedBERT's BIOSSES correlation from 0.633 to 0.828 and its within-versus-across-domain separation from 1.05× to 1.63×. A second pass we call BODHI — which mines hard negatives from edges that are absent in a biomedical knowledge graph — pushes separation to 2.30× and the discrimination gap to +0.392, costing 4.5% on BIOSSES. Then we make it run. On an Intel Xeon 6737P with AMX, OpenVINO cuts single-query latency from 1,367 ms to 10 ms (133×) and reaches 555 sentences per second. One result contradicts standard deployment advice: FP16 beats INT8 on this silicon at every batch size we serve, and we explain why. The same model on a no-AMX Ice Lake instance runs 13–27× slower.

We release the benchmark suite, the training corpora, the BODHI hard-negative generator, and the OpenVINO conversion scripts. The LBM models the human. This paper builds the embedding layer that lets it do so without inventing connections that were never there.

**Keywords —** behavioural foundation models, biomedical NLP, embedding anisotropy, cross-domain causal discovery, ontology-graph fine-tuning, Matryoshka representations, Intel AMX BF16, OpenVINO, contrastive learning, Large Behavioural Model

## 1. The problem, stated plainly

A person is not a paragraph. The models that dominate AI today are built around language: the unit is the token, the objective is to predict the next one, and success is measured in fluency. They are good at text. That does not make them good at people. People are continuous, stateful, and causally tangled — biology feeding cognition feeding mood feeding behaviour, across timescales from minutes to decades. Modelling a person needs a different primitive than the token, a different objective than next-token prediction, and a different memory than a context window.

We call that primitive a Large Behavioural Model. An LBM is a foundation model whose subject is a human, represented across clinical, genomic, psychological, behavioural, environmental, and physiological signals and across time. It is not a chatbot and it is not a competitor to language models. The two sit side by side and do different jobs. Language models model language; an LBM models the person the language is about. The architecture is described in our companion paper (arXiv:2605.27580). This paper is about one component of it, and about a failure that breaks the whole thing if you get it wrong.

The LBM keeps a graph of each user's life. Nodes are events — a lab result, a journal entry, a counselling note, a wearable reading. Edges are relationships: this caused that, this preceded that, this is a recurrence of that. The model grows the graph by walking it, and one of the signals it walks on is embedding proximity. When two events have no edge between them but their embeddings sit close together, the model treats that closeness as a reason to ask whether a causal link exists, and if the answer is yes, it draws the edge. Over months, this is how the LBM turns a pile of disconnected records into a causal map of a single life.

Now the failure becomes obvious. The whole loop trusts the geometry. If the embedding space puts unrelated things close together, the model hypothesises edges that do not exist, and the user's causal map fills up with fiction. If it puts genuinely linked things far apart, the model never connects them and the map has holes. Either way, the reasoning that runs on top of the graph is reasoning on a lie. So the question this paper answers is narrow and load-bearing: can an embedding model place causally related events close together and causally unrelated events far apart, even when those events come from different domains and use completely different vocabularies? The honest starting answer, for every model you can download today, is no.

Here is what we found and what we did about it:

- Off-the-shelf biomedical encoders fail the one test that matters for this use case. On cross-domain discrimination they score 0% accuracy, with unrelated pairs landing at 0.76–0.92 cosine. The cause is the well-known anisotropy of BERT embeddings.
- A two-pass contrastive fine-tune fixes it, and we measured the result rather than guessing. The second pass, BODHI, learns from edges that are missing in a biomedical knowledge graph and produces the cleanest separation we have seen: intra/inter-domain ratio 2.30×, discrimination gap +0.392, F1 of 93.0%.
- It runs on commodity CPUs. On Intel Xeon with AMX, OpenVINO delivers a 133× latency cut and 555 sentences/sec, no GPU required. FP16 beats INT8 here, which is the opposite of the usual advice, and we show why.

- AMX silicon matters. The identical model on a no-AMX Ice Lake server runs 13–27× slower, and the INT8 quantisation that Ice Lake needs for speed rounds away half to four-fifths of the discrimination signal we worked to create.

## 2. What an LBM needs from its embeddings

Retrieval systems use embeddings as a similarity index. You ask a question, you get back the documents that look most like it, and a language model decides what to do with them. If the index returns something slightly off, the language model shrugs it off. The cost of an embedding error is a worse search result, and the system absorbs it.

Graph reasoning has no such cushion. The consumer of the embedding is not a forgiving language model; it is a traversal routine that treats proximity as evidence and writes that evidence into a permanent graph. A wrong edge does not get re-ranked away. It sits in the graph and corrupts every future walk that passes through it. Errors compound instead of washing out. That single difference is why an embedding space that is fine for search can be useless here.

What the LBM actually needs is not similarity but causal coherence that survives a change of vocabulary. Four behaviours separate the two. The model has to pull together events that sit on the same causal chain even when they are written in different languages — a journal line that reads “slept four hours, anxious all morning” belongs next to a lab value of “cortisol 28 µg/dL,” because sleep loss drives the cortisol drives the anxiety, but one is a feeling and the other is a number. It has to push apart events that merely share surface words — “elevated cortisol” and “market volatility” both invoke stress, but there is no mechanism connecting them, so they must end up far apart. It has to link across vocabularies that look nothing alike — the genetics node “BRCA1 pathogenic variant” and the clinical node “increased breast-cancer screening” are cause and consequence, one molecular and one procedural. And it has to recognise the same underlying state described two different ways months apart, so the model can see a recurring episode rather than two unrelated entries.

A model trained to predict masked tokens over PubMed learns none of this directly. It learns that “cortisol” and “stress” co-occur. It never learns that sleep deprivation causes elevated cortisol causes anxiety, because that chain runs across three domains and three vocabularies and no masked-token objective rewards stitching them together. Left alone, the model defaults to surface co-occurrence, which is exactly the wrong signal: it rates unrelated-but-similar-sounding pairs as close, and that is the failure we measured.

## 3. Models, benchmarks, and hardware

### 3.1 The three encoders

We evaluated three biomedical encoders that between them cover the architectures people actually deploy: a domain-continued BERT (BioBERT), a from-scratch domain-pretrained BERT (PubMedBERT), and a discriminator-trained model (BioM-ELECTRA-Large). We fine-tuned all three. We kept ELECTRA in the comparison even though it loses, because the way it loses is informative.

| Model | Backbone | Pretraining corpus | Dim | Objective |
|---|---|---|---|---|
| BioBERT v1.1 (dmis-lab) | BERT-Base | PubMed + PMC full text | 768 | MLM + NSP |
| PubMedBERT (Microsoft) | BERT-Base | PubMed, from scratch | 768 | Domain MLM |
| BioM-ELECTRA-Large (sultan) | ELECTRA-Large | Biomedical, SQuAD2 fine-tune | 1024 | Replaced-token detection |

*Table 1. The three encoders. Production candidates are the two BERT-Base models; ELECTRA is a contrast case.*

## 3.2 Six tests

We built six diagnostics, each aimed at one thing the embedding has to get right for graph reasoning to work. We score them individually so a model can pass one and fail another, which is exactly what happens.

| Test | Measures | Why it matters |
|---|---|---|
| B1 Within-domain | Cosine between related pairs in one domain | Same-domain events must cluster, or intra-domain chains break |
| B2 Cross-domain | Cosine between unrelated pairs across domains | The one that matters: stops the LBM drawing false cross-domain edges |
| B3 STS (BIOSSES) | Rank correlation with clinician similarity scores | Checks that proximity tracks expert judgement |
| B4 Throughput / latency | Sentences/sec across batch sizes; p50/p95 latency | Sets how fast the graph can ingest new events |
| B5 Hard negatives | Accuracy on anchor / positive / hard-negative triplets | Catches the deceptively-similar-but-unrelated case |
| B6 Geometry | Inter- over intra-domain distance ratio | Above 1.0, the space is globally usable as evidence |

*Table 2. The six diagnostic suites.*

## 3.3 Two machines

We ran on two Intel platforms that differ in one decisive way. The primary machine is an Intel Xeon 6737P (Granite Rapids) with AMX, the tile hardware that does BF16 matrix multiply natively. The reference machine is an AWS c6i.8xlarge (Ice Lake) with no AMX at all. Holding the software stack, the model code, and the calibration data constant across the two isolates exactly what the AMX silicon buys. Intel India provided access to the Granite Rapids server under NDA; we thank their team.

| | Primary: Xeon 6737P (Granite Rapids) | Reference: c6i.8xlarge (Ice Lake) |
|---|---|---|
| Generation | 4th-Gen Xeon | 3rd-Gen Xeon |
| Physical cores | 64 (2 × 32) | 16 |
| Logical cores | 128 | 32 |
| L3 cache | 288 MiB (144 per NUMA node) | 54 MiB |
| Memory | 1 TB DDR5-6400 | 61 GB DDR4 |

| | Primary: Xeon 6737P (Granite Rapids) | Reference: c6i.8xlarge (Ice Lake) |
|---|---|---|
| AMX BF16 | Yes, native tiles | No |
| AVX-512 VNNI | Yes | Yes (INT8 only) |
| OS / kernel | Ubuntu 24.04 / 6.8.0 | Ubuntu 24.04 / 6.17.0-aws |
| OpenVINO / PyTorch | 2026.1 / 2.11 | 2026.1 / 2.11 |

*Table 3. The two platforms. Everything but the silicon is held constant, so the gap between them is the AMX gap.*

On the AMX server we pinned every worker to NUMA-balanced cores (numactl, KMP_AFFINITY=compact, OMP_NUM_THREADS=128) and told OpenVINO to use BF16 (INFERENCE_PRECISION_HINT=bf16). We checked that AMX was actually engaged through OPTIMIZATION_CAPABILITIES. On the c6i the same BF16 hint is silently ignored — its capability list has no BF16 entry — so anything labelled BF16 there runs as FP32. That detail turns out to matter.

## 4. How badly the pretrained models fail

Before any fine-tuning, this is what you get when you pull these models off the hub and embed your data. We start with the headline numbers, then look at the one test that sinks all three.

| Metric | BioBERT | PubMedBERT | ELECTRA |
|---|---|---|---|
| Within-domain cosine (avg) | 0.882 | 0.946 | 0.958 |
| Cross-domain cosine (avg, lower better) | 0.756 | 0.910 | 0.917 |
| BIOSSES ρ | 0.762 | 0.633 | 0.289 |
| Hard-negative accuracy | 80% | 60% | 80% |
| `Geometry ratio (target ≥ 1)` | 0.652 | 0.581 | 0.504 |
| Discrimination gap | +0.051 | +0.013 | +0.007 |

*Table 4. Pre-finetune scores. ELECTRA wins on raw within-domain similarity and loses everywhere that counts for the LBM. BioBERT is the least-bad baseline.*

There is a trap in that table. ELECTRA has the highest within-domain similarity, 0.958, and if you were grading on that alone you would pick it. But high within-domain similarity with high cross-domain similarity is not discrimination — it is a model that thinks everything is similar to everything. ELECTRA also has the worst BIOSSES (0.289) and the worst geometry (0.504), because its outputs pile up in a narrow band and lose all resolution. BioBERT, with weaker raw numbers, is the only model that begins to separate signal from noise, so it is our reference baseline from here on.

### 4.1 Ten pairs

The quickest way to see the failure is a ten-pair test: eight pairs that should score high, two that should score low. Everyone passes the eight. Nobody passes the two.

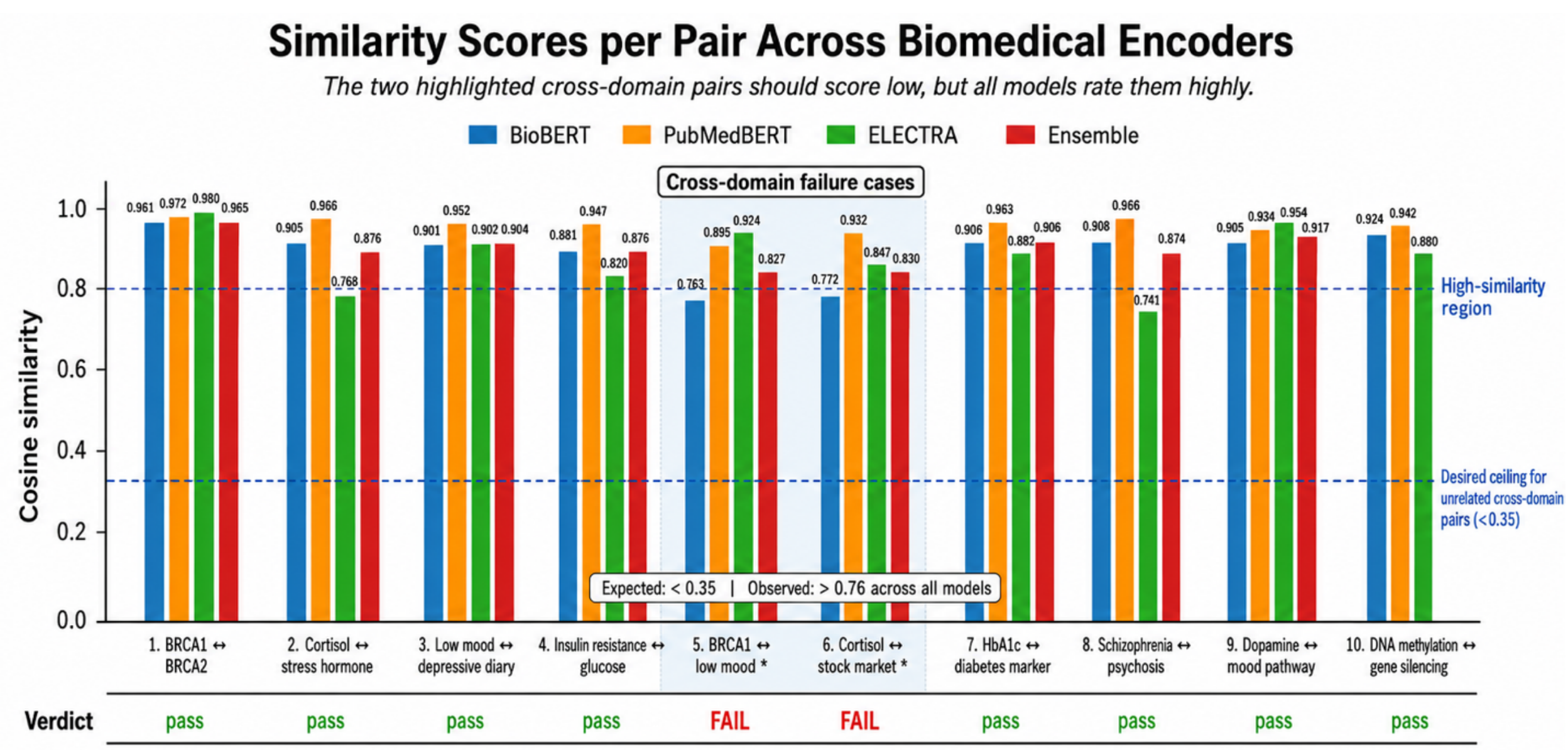

*Figure 1. Per-pair cosine across the three models and their ensemble. The two pink columns are cross-domain pairs that should score near zero; every model rates them above 0.76. "BRCA1 vs low mood" and "cortisol vs stock market" are the failures.*

| Pair | BioBERT | PubMedBERT | ELECTRA | Ensemble | Verdict |
|---|---|---|---|---|---|
| BRCA1 ↔ BRCA2 | 0.961 | 0.972 | 0.980 | 0.965 | pass |
| Cortisol ↔ stress hormone | 0.905 | 0.966 | 0.768 | 0.876 | pass |
| Low mood ↔ depressive diary | 0.901 | 0.952 | 0.902 | 0.904 | pass |
| Insulin resistance ↔ glucose | 0.881 | 0.947 | 0.820 | 0.876 | pass |
| BRCA1 ↔ low mood * | 0.762 | 0.895 | 0.924 | 0.827 | FAIL |
| Cortisol ↔ stock market * | 0.773 | 0.932 | 0.847 | 0.830 | FAIL |
| HbA1c ↔ diabetes marker | 0.906 | 0.963 | 0.882 | 0.906 | pass |
| Schizophrenia ↔ psychosis | 0.908 | 0.966 | 0.747 | 0.874 | pass |
| Dopamine ↔ mood pathway | 0.905 | 0.934 | 0.954 | 0.917 | pass |
| DNA methylation ↔ gene silencing | 0.814 | 0.924 | 0.942 | 0.880 | pass |

*Table 5. The ten-pair test. Cross-domain pairs (*) should fall below 0.35 and instead clear 0.82.*

## 4.2 The test that sinks everything

B2 measures the average cosine the models assign to pairs that are genuinely unrelated and happen to live in different domains. Lower is better; the target is below 0.35. The result is not close.

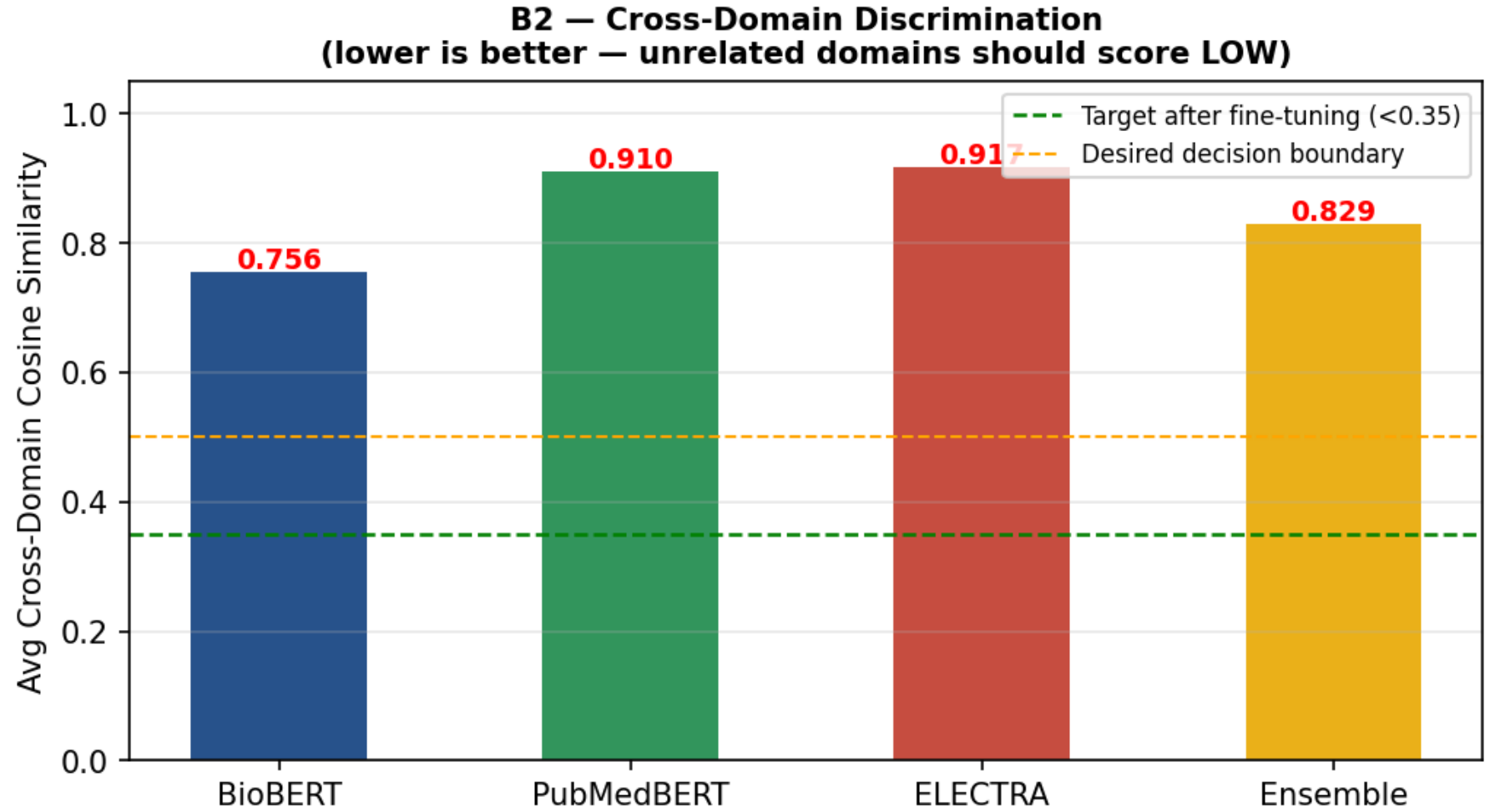

*Figure 2. Cross-domain discrimination (B2). Every model clears 0.75 where it should sit below 0.35. BioBERT at 0.756 is the least-bad; PubMedBERT and ELECTRA are worse. Accuracy on this test is 0% for all three.*

A model that scores cortisol against a finance headline at 0.83 cannot be trusted to decide which events in a person's life are connected. The discrimination margin — the gap between the noise floor of unrelated pairs and the signal ceiling of real links — is far too thin to set a threshold on. Figure 3 confirms the within-domain side is fine, which rules out the easy explanation that the models are simply weak. They are not. They are anisotropic.

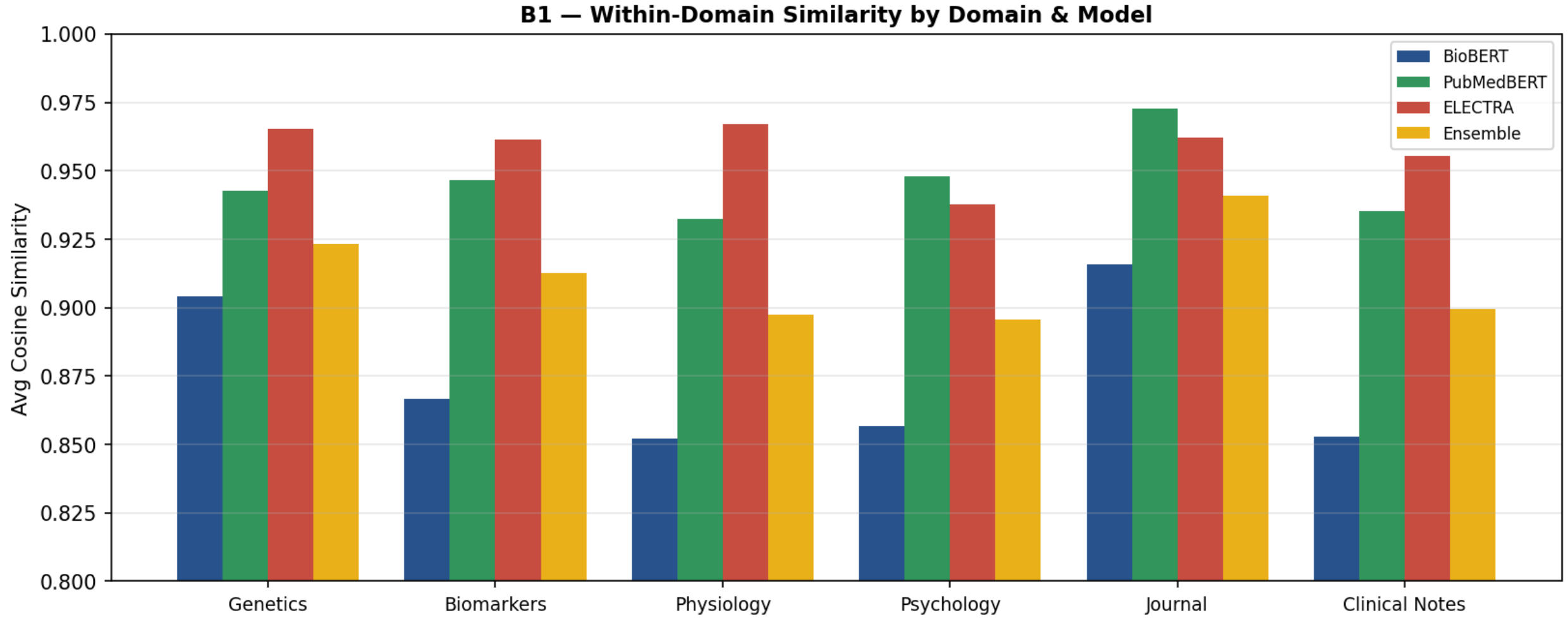

*Figure 3. Within-domain similarity (B1) by domain. All three models cluster same-domain content tightly. The problem is not within domains; it is between them.*

## 5. Why they fail: anisotropy

The pattern across B2, B3, and B6 is the signature of a problem the embedding literature has documented for years. BERT-family sentence vectors do not spread out over the unit sphere. They collapse into a narrow cone, all pointing roughly the same way, clustered around the corpus mean. The practical consequence is that the cosine between any two random sentences sits well above zero. Unrelated content reads as 0.4 to 0.7 similar simply because every vector lives in the same small region of space.

Search tolerates this. A language model reads the top results and ignores the junk, so a high noise floor costs a little precision and nothing else. Graph reasoning does not get that luxury. The LBM has to put its edge-hypothesis threshold somewhere on the cosine axis, and if the noise floor is 0.75 while real links sit at 0.95, there is no threshold that separates them across millions of nodes. Whitening, contrastive sentence training, and prompt-based pooling have all been shown to reduce anisotropy on standard similarity benchmarks. None of them have been tested on the thing we care about, which is whether the space pulls cross-domain causal pairs together while pushing cross-domain look-alikes apart. So we built training that targets that directly.

## 6. The fix: two passes, and one good idea

### 6.1 Pass 1 — contrastive training on 72,034 pairs

Pass 1 fine-tunes each model on 72,034 (anchor, positive, hard-negative) triplets drawn from eight public datasets and topped up with synthetic cross-domain hard negatives. The loss is Matryoshka Loss wrapped around Multiple Negatives Ranking Loss. MNRL turns every other example in a batch of 128 into a negative, so each anchor is pushed away from 127 others at once, plus its explicit hard negative. The Matryoshka wrapper trains five nested dimensions at the same time — 768, 512, 256, 128, 64 — which means one model serves the full 768-d vector for graph storage, a 256-d vector for fast proximity scans during traversal, and a 64-d vector for an on-device LBM, with no retraining.

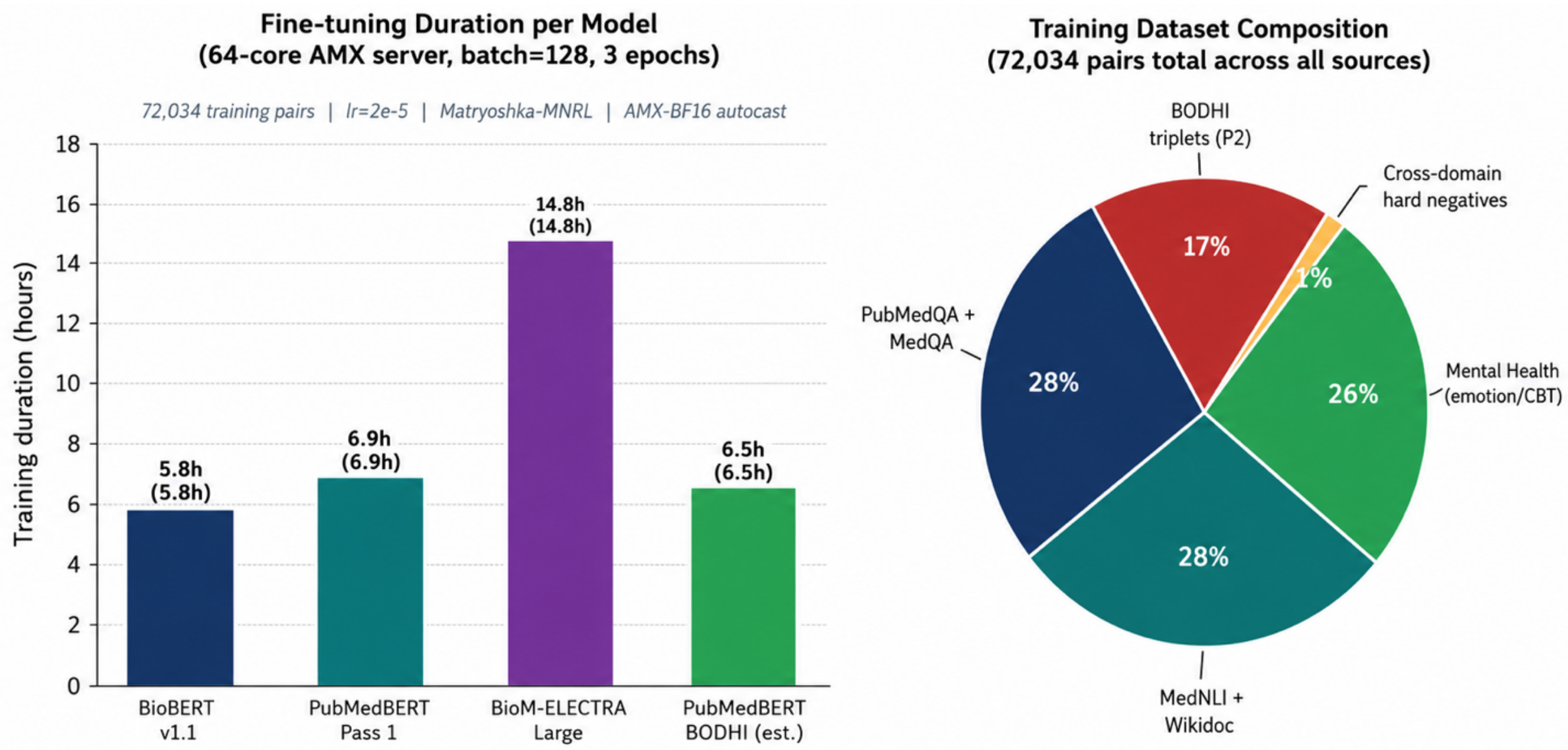

*Figure 4. Pass 1 training. Left: wall-clock time per model on 64 reserved AMX cores — BioBERT 5.8 h, PubMedBERT 6.9 h, ELECTRA 14.8 h. Right: the eight-source mix of biomedical, clinical, and psychology data behind the 72,034 pairs.*

| Dataset | Domain | Role |
|---|---|---|
| all-nli | general | entailment pairs → hard negatives |
| BIOSSES | biomedical STS | gold scores → calibration |

| Dataset | Domain | Role |
|---|---|---|
| medical_questions_pairs | clinical Q&A | expert pairs → positives |
| MedNLI (MIMIC-III) | clinical NLI | entailment / contradiction |
| PubMedQA | biomedical QA | question + abstract → anchor/positive |
| emotion | psychology | within-domain positives |
| go_emotions | psychology | fine-grained emotion pairs |
| mental_health_counseling | psychology / journal | dialogue → journal-style anchors |

*Table 6. The eight Pass-1 sources. Cross-domain hard negatives are generated by pairing biomedical anchors with psychology / journal text that shares surface words but no mechanism.*

## 6.2 Pass 2 — BODHI, learning from what is missing

Pass 1 helps, but its hard negatives are only as good as the heuristics that made them, and lexical-overlap heuristics miss the point of the failure. The models do not confuse things that look alike on the surface; they confuse things that are conceptually unrelated. What we needed was a source of hard negatives that encodes whether two concepts are actually connected, judged by something other than word overlap.

A knowledge graph is exactly that source. In a curated biomedical graph, an edge between two concepts is a domain expert's assertion that they are linked, and the absence of an edge is an assertion that they are not. We use the BODHI biomedical knowledge graph this way. Every explicit edge of type PRESENT_IN, CHILD_OF, or IMPACTS becomes a positive pair. Every pair of concepts that are both in the graph but have no edge between them — no path of length two or less — becomes a hard negative. So "BRCA1 → hereditary cancer syndrome" is a positive, because the edge exists, and "BRCA1 vs depression" is a hard negative, because no expert ever connected them, however similar a pretrained model might think they are. Pass 2 takes the Pass-1 PubMedBERT checkpoint and trains two more epochs on roughly 3,500 of these ontology triplets, producing a distinct model: PubMedBERT BODHI.

There is a symmetry worth naming. BODHI trains the embedding on absent edges in one graph so that, in deployment, the embedding can help the LBM hypothesise absent edges in another — the user's personal causal graph. We train the model to do the job it will be asked to do. That is the whole idea, and it works better than anything else we tried.

| Setting | Value |
|---|---|
| Loss | MatryoshkaLoss(MNRL), dims [768, 512, 256, 128, 64] |
| Batch | 128 (127 in-batch negatives) + explicit hard negatives |
| Learning rate | 2e-5 BERT models, 5e-6 ELECTRA |
| Epochs | 3 (Pass 1) + 2 (BODHI) |

| Setting | Value |
|---|---|
| Precision | BF16 autocast on AMX |
| Cores | 64 reserved (one NUMA node), inference protected on the other |

*Table 7. Training configuration. BODHI is a second pass on the Pass-1 checkpoint, not a separate model from scratch.*

## 7. What the fine-tuning bought

These are measured results, not projections. The geometry that was collapsed before training comes apart into clean, separated clusters after it.

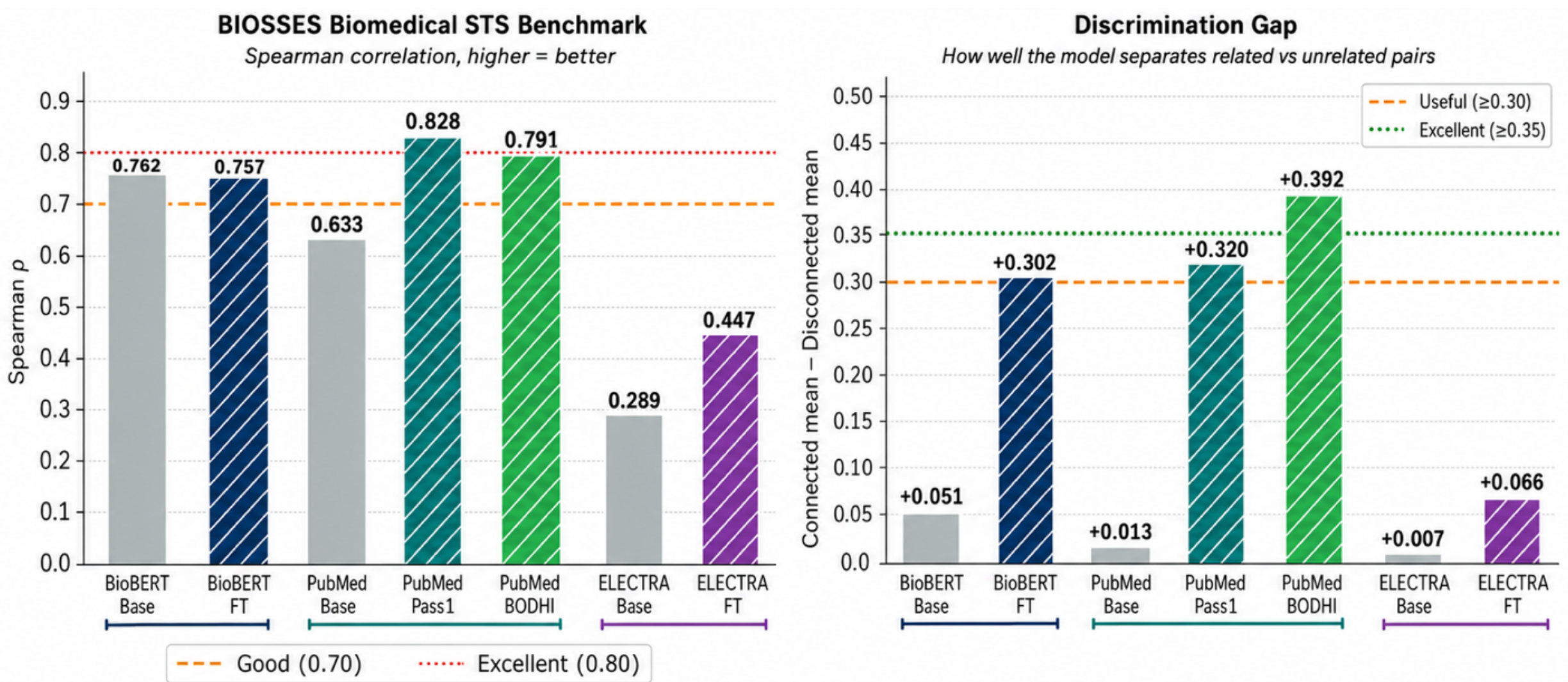

*Figure 5. Left: BIOSSES rank correlation. PubMedBERT Pass 1 hits 0.828, past the 0.80 line that counts as strong clinician-level agreement. ELECTRA never gets close, even after fine-tuning. Right: the discrimination gap. BODHI reaches +0.392, the widest separation in the study and 30× the pretrained baseline.*

PubMedBERT Pass 1 lands at 0.828 BIOSSES, comfortably into strong-agreement territory. BioBERT loses a hair of BIOSSES in fine-tuning (0.762 to 0.757) but trades it for a discrimination gap that jumps from +0.051 to +0.302, which is the trade you want. ELECTRA climbs from 0.289 to 0.447 and stays unusable, because a model pretrained for span-extraction QA does not reshape into a good embedder no matter how much biomedical data you show it. That is the informative loss we kept ELECTRA around to demonstrate.

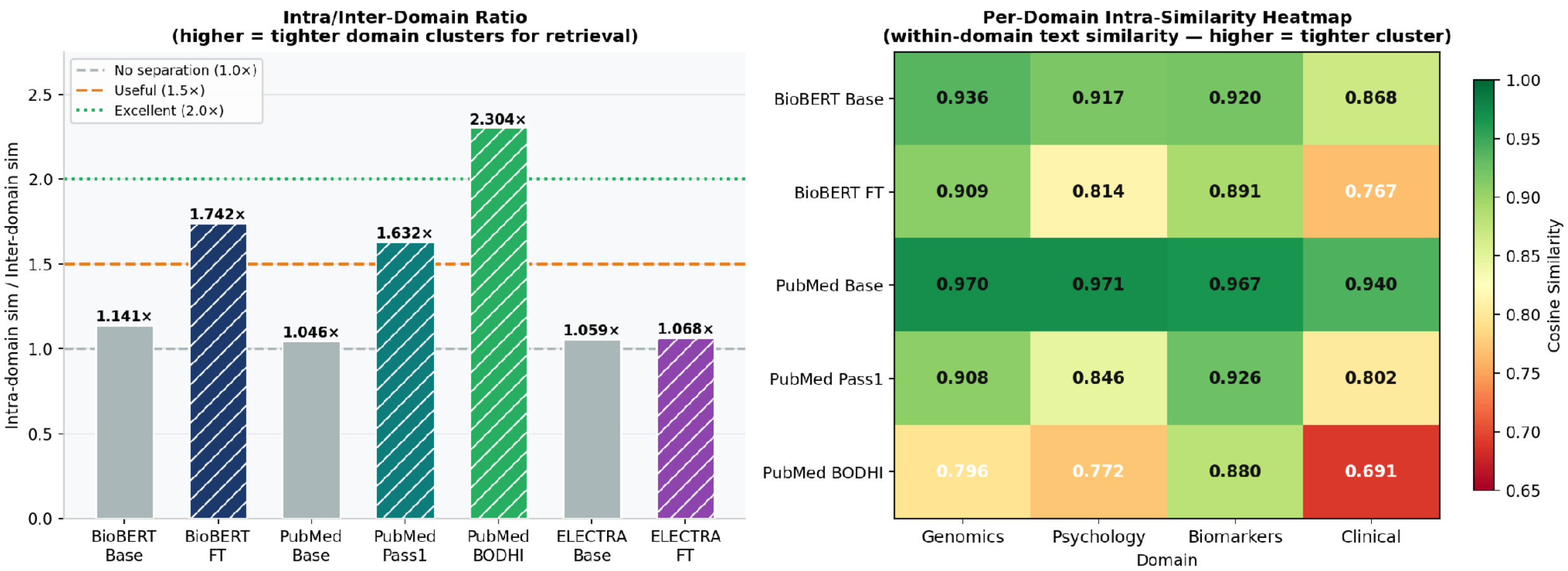

*Figure 6. Left: the intra/inter-domain ratio climbing from 1.05× (collapsed) through 1.63× (Pass 1) to 2.30× (BODHI) — the Pass 1 to BODHI step alone is +41% separation. Right: the per-domain similarity heatmap. BODHI does most of its work by driving down cross-domain similarity, not by tightening within-domain clusters.*

The geometry ratio is the number that predicts whether the LBM can use proximity at all. Below 1.0 the space is useless; same-domain pairs are no closer than cross-domain pairs. Above 1.5 it becomes a reliable signal. Pass 1 reaches 1.63× and BODHI reaches 2.30×. The way BODHI gets there matters: it suppresses inter-domain similarity from 0.533 to 0.341 while barely touching the within-domain numbers, which is precisely the move that stops false cross-domain edges without breaking real within-domain chains.

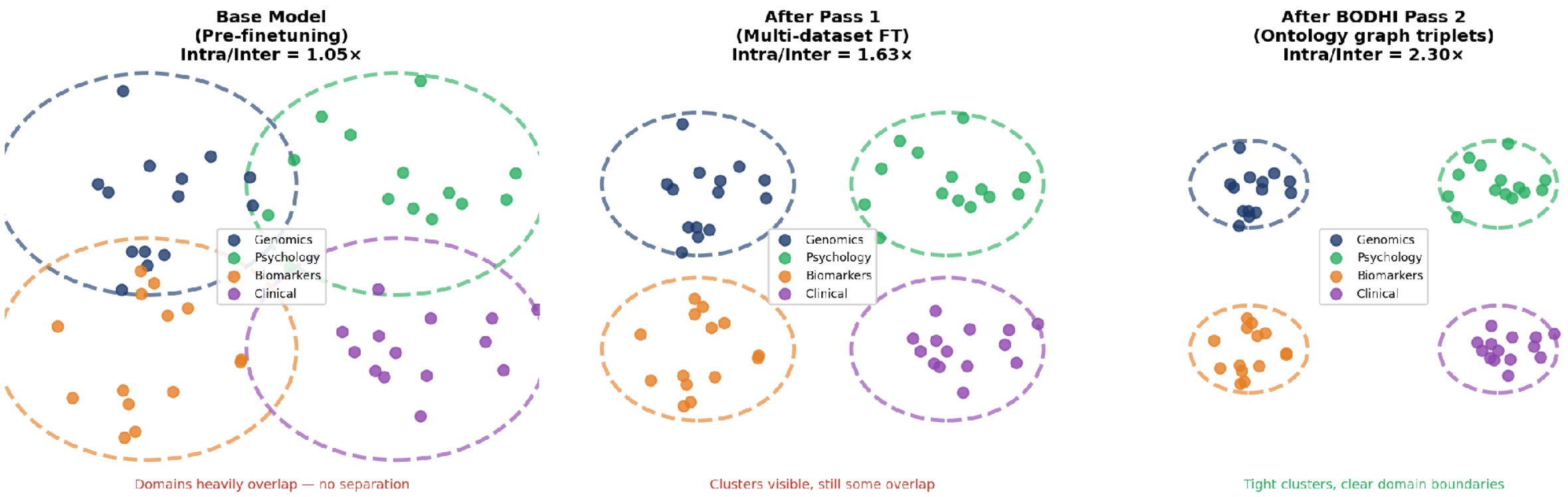

*Figure 7. The same story as a picture. Base model: three domains in one overlapping blob (1.05×). After Pass 1: clusters appear but bleed into each other (1.63×). After BODHI: tight, cleanly separated clusters (2.30×). Schematic; the real space is 768-dimensional.*

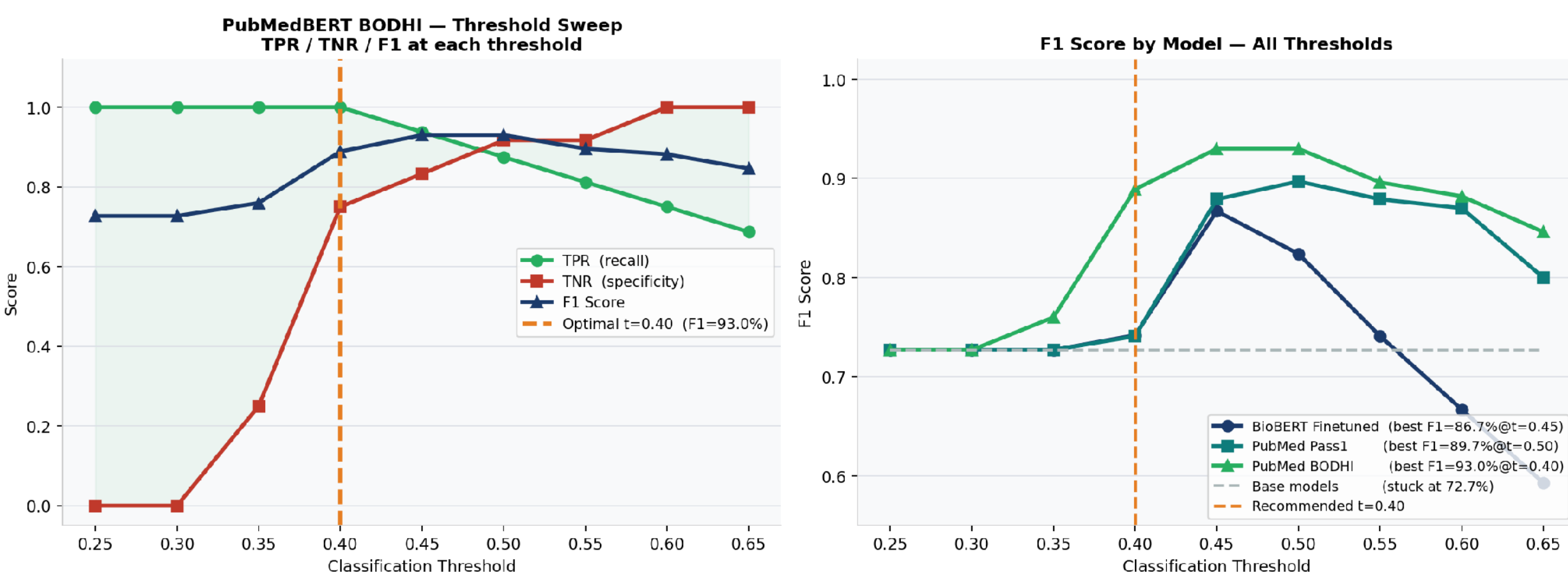

*Figure 8. Picking the decision threshold. Left: the BODHI sweep peaks at F1 = 93.0% at τ = 0.40. Right: best F1 by stage. Pretrained models top out at 72.7% — they cannot separate at any threshold. Pass 1 reaches 89.7%, BODHI 93.0%.*

The intrinsic geometry shows up as an operational number through the threshold sweep. Set one global cosine cutoff, call anything above it a link and anything below it not, and read off the best F1. Pretrained models are stuck at 72.7% no matter where you put the threshold, because there is no threshold that works on a collapsed space. Pass 1 reaches 89.7% at τ = 0.50. BODHI reaches 93.0% at τ = 0.40. Those two thresholds are the deployment parameters the LBM uses to trade precision against recall on edge hypotheses.

# 8. Making it fast: AMX, OpenVINO, and a surprise

## 8.1 133× on commodity silicon

A correct embedding that takes 1.4 seconds per query is not deployable for a system that ingests a stream of life events. After fine-tuning we export each model to OpenVINO with FP16 weights and a BF16 runtime, using the one flag that keeps the full sequence output instead of a task head:

```
optimum-cli export openvino \
    --model <hf_model_id> \
    --task feature-extraction \
    --weight-format fp16 --group-size 64 --ratio 1.0 \
    <output_dir>/
```

On the same hardware, with the same code, this turns a 1,367 ms PyTorch query into a 10 ms one. Same accuracy — the cosine scores move by less than 0.0005. The throughput side scales the same way, with PubMedBERT topping out at 617.8 sentences per second at batch 256 and 555 on the fine-tuned model at batch 512.

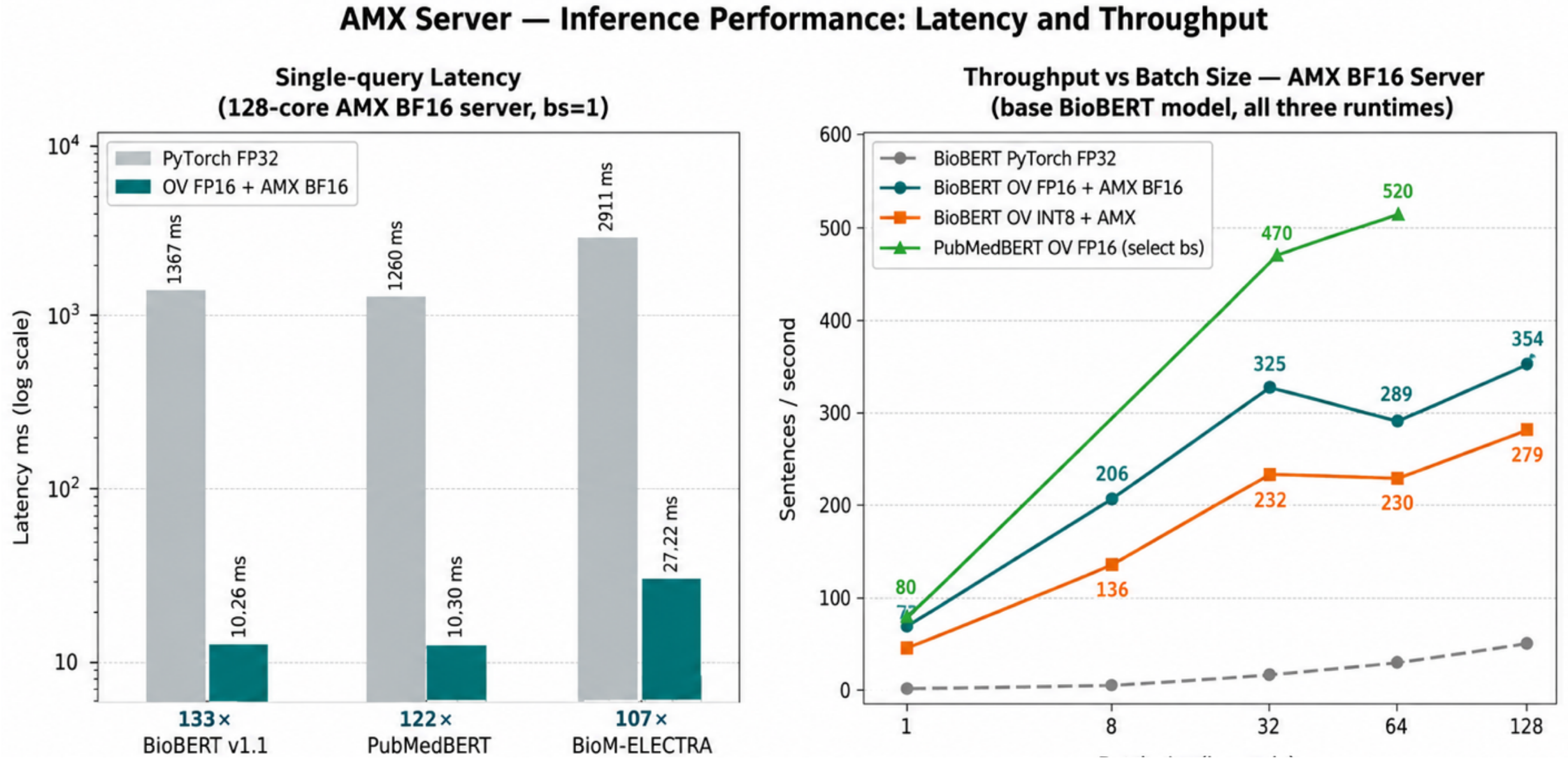

*Figure 9. Left: single-query latency, log scale. BioBERT 1366.6 → 10.26 ms (133×), PubMedBERT 1259.7 → 10.30 ms (122×), ELECTRA 2911.3 → 27.22 ms (107×). Right: throughput versus batch size — OpenVINO FP16 dominates the PyTorch baseline everywhere.*

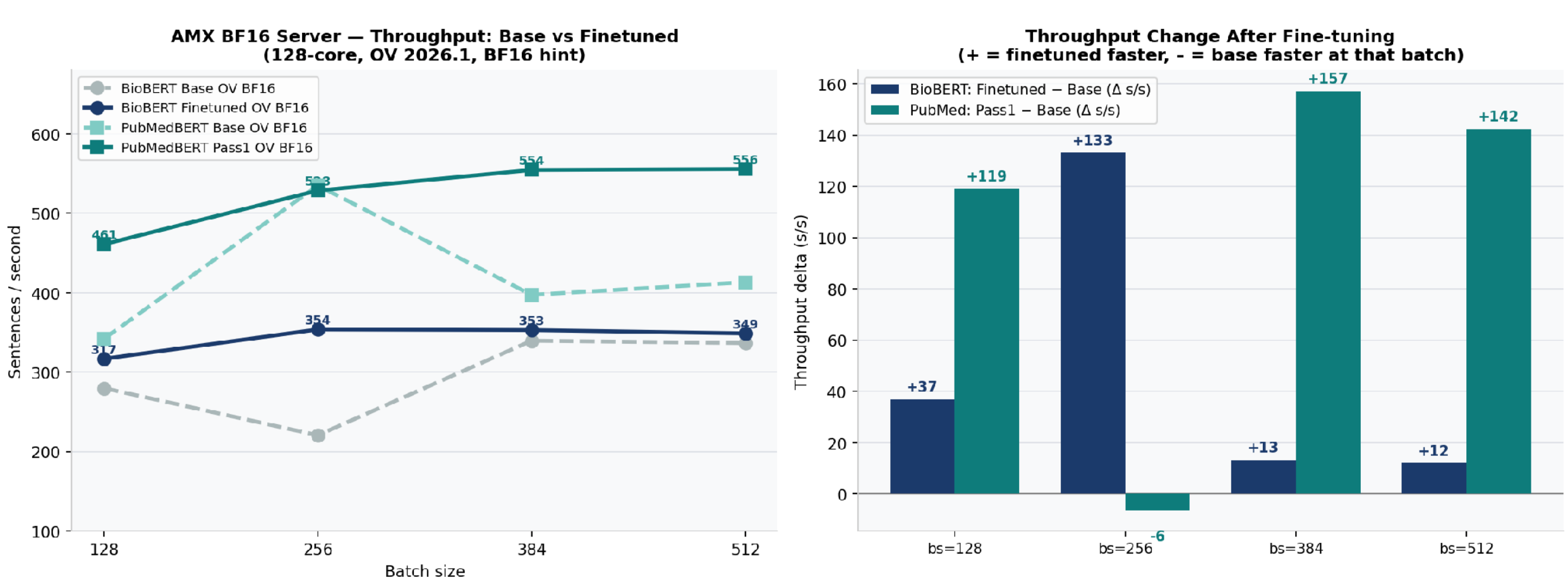

*Figure 10. Fine-tuning changes weights, not architecture, so it does not cost throughput. The fine-tuned PubMedBERT is actually faster than the base model at large batches — up to +157 sentences/sec.*

## 8.2 FP16 beats INT8, which is not supposed to happen

Standard deployment advice says quantise to INT8 for inference. On this AMX silicon that advice is wrong, and we have the curves to prove it. For every batch size up to 256, and for all three models, OpenVINO FP16 with a BF16 runtime is faster than OpenVINO INT8.

The reason is specific to how AMX handles the two formats. AMX has a BF16 tile unit and an INT8 tile unit, and on paper the INT8 unit does twice the work per cycle. But the BF16 weights are stored as FP16 on disk and cast to BF16 at load, which is free on AMX — no kernel runs. INT8 weights have to be dequantised back to a wider format before they can feed the residual adds in attention and the feed-forward block, and that dequantisation runs on scalar units, not the tile, adding cost to every forward pass. The crossover where INT8's extra tile throughput finally pays for its dequantisation overhead sits around batch 512 for BERT-base and higher for BERT-large. Production serving lives below that. So FP16 wins on speed, and because BF16 carries the same exponent range as FP32, it wins on accuracy too. The practical rule for this hardware class: do not quantise BERT-class embedders to INT8 at serving batch sizes. Use FP16 weights with a BF16 runtime.

This is not an abstract preference. The accuracy evidence across both machines shows two quantisation paths that should never ship: PyTorch dynamic INT8, whose cosine fidelity collapses to 0.68–0.94, and OpenVINO INT4, which mauls PubMedBERT down to 0.90. FP16 and BF16 stay lossless; OpenVINO INT8 holds 0.995 fidelity but, as the next section shows, pays for it on the discrimination gap.

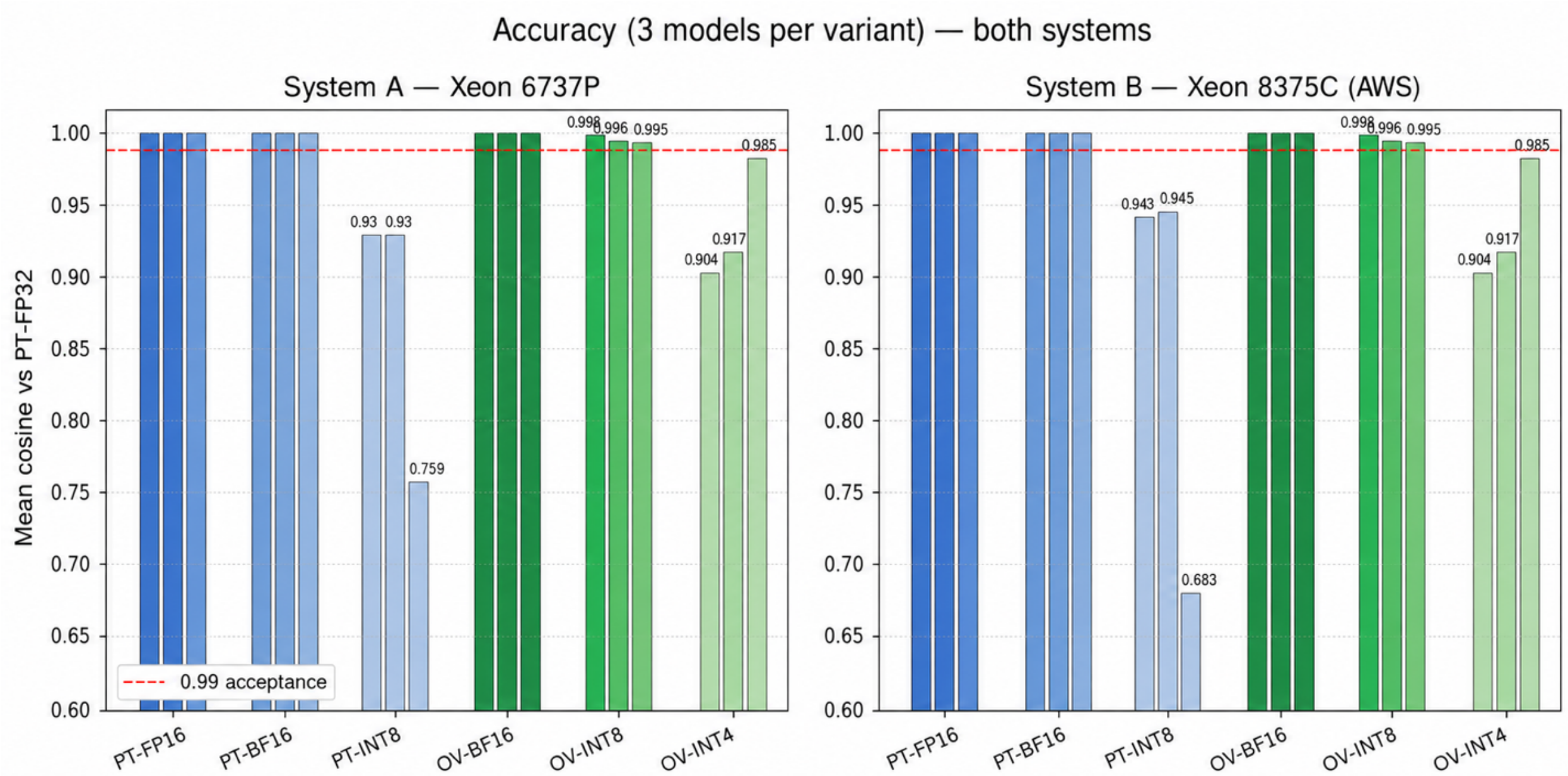

*Figure 11. Embedding fidelity (cosine vs FP32 reference) across six precisions on both machines. FP16 and BF16 are lossless. OpenVINO INT8 holds 0.995. PyTorch dynamic INT8 (0.68–0.94) and OpenVINO INT4 (0.90) both fail — and the PyTorch INT8 failure reproduces on both machines, so it is the method, not the silicon.*

## 8.3 The AMX gap

To separate the silicon from the software, we ran the identical stack on the no-AMX Ice Lake c6i. The AMX server is 13 to 27× faster at batch 256. ELECTRA shows the widest gap, 27×, because the larger model has more matrix multiply to feed the tiles.

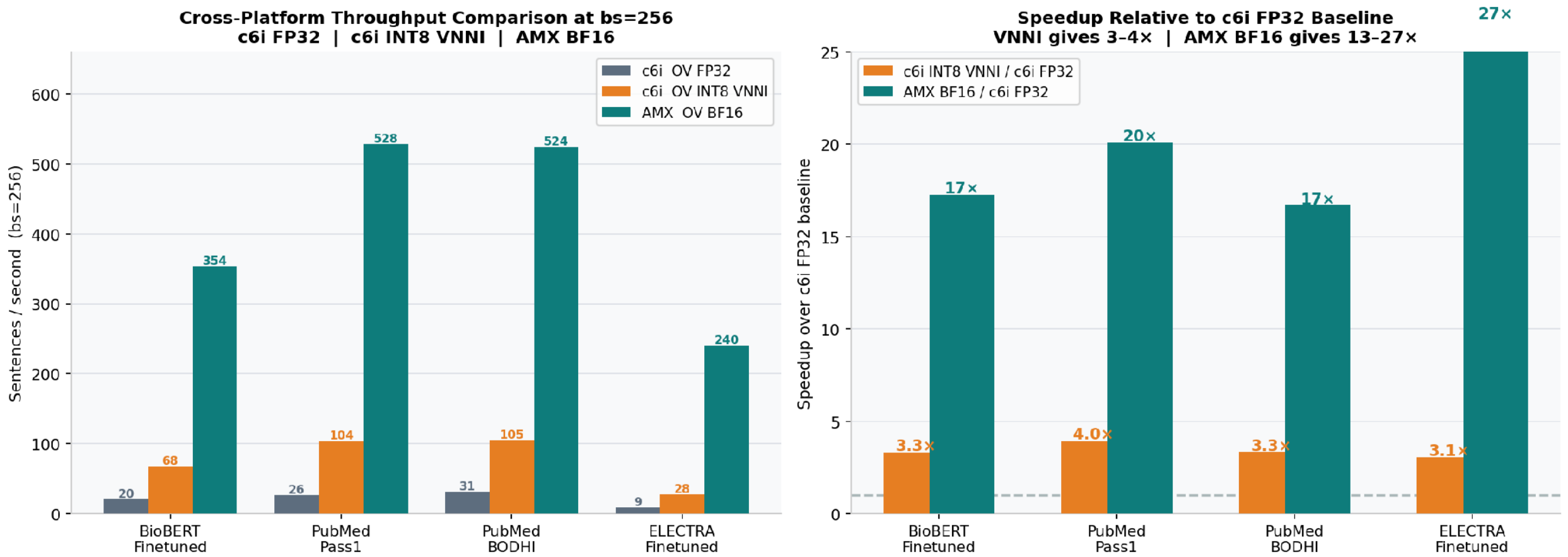

*Figure 12. Cross-platform throughput at batch 256. Left: absolute sentences/sec across c6i FP32, c6i VNNI INT8, and Xeon AMX BF16. Right: speedup over the c6i FP32 baseline — AMX BF16 is 13–27× faster, and 5–9× faster than the best the c6i can manage with INT8.*

Ice Lake can claw back some speed with AVX-512 VNNI INT8, which is genuinely 3–4× faster than FP32 there. But that speed comes out of the discrimination gap. INT8 rounding flattens exactly the signal the fine-tuning produced: BioBERT keeps only 17% of its gap, PubMedBERT Pass 1 keeps 50%, BODHI 36%. A common early mistake makes it worse — quantising only the feed-forward layers to "protect accuracy" leaves 85% of the compute in FP32 and produces no speedup at all. You have to quantise everything to reach the VNNI units, and quantising everything is what destroys the gap. On AMX you simply do not face this trade, which is the real argument for the hardware.

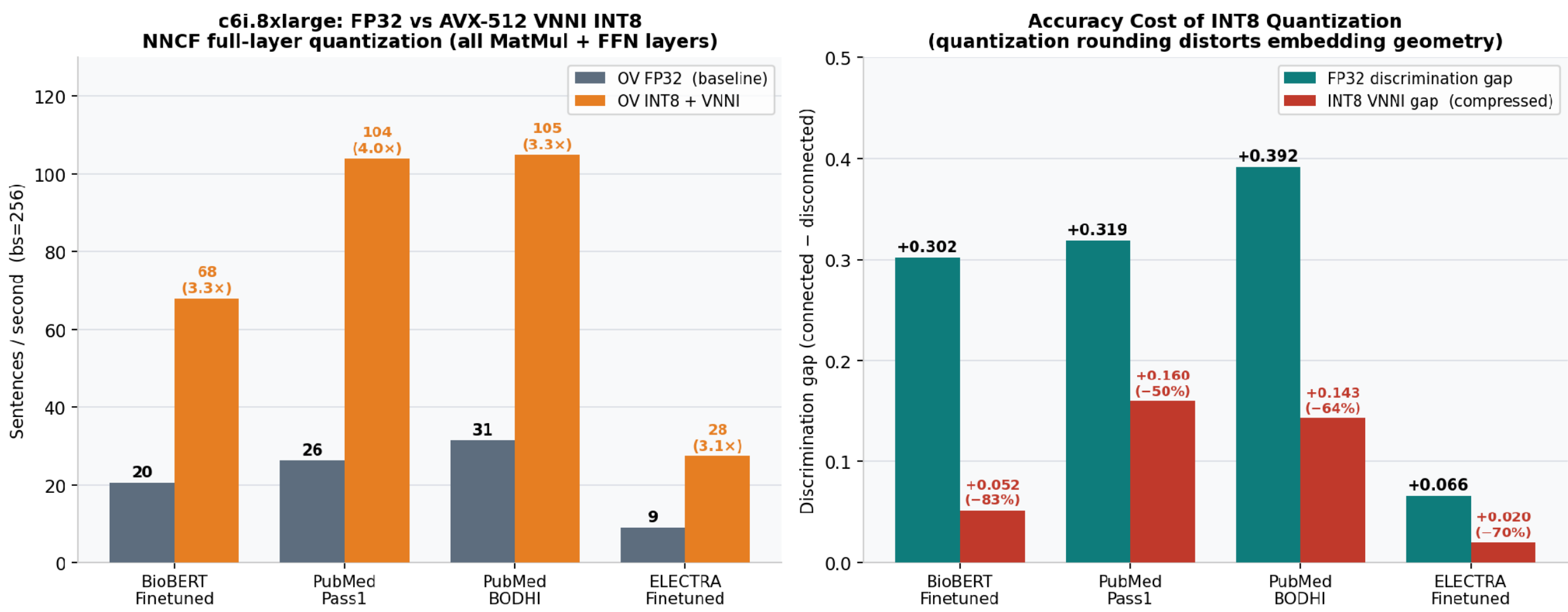

*Figure 13. The c6i INT8 trade-off. Left: VNNI INT8 is 3–4× faster than FP32. Right: it keeps only 17–50% of the discrimination gap. Speed on Ice Lake costs the very thing the fine-tuning created.*

## 9. Production

Two regimes matter in deployment: batch ingestion, where you embed a backlog as fast as possible, and online serving, where many clients hit the model at once. We characterised both on the AMX server.

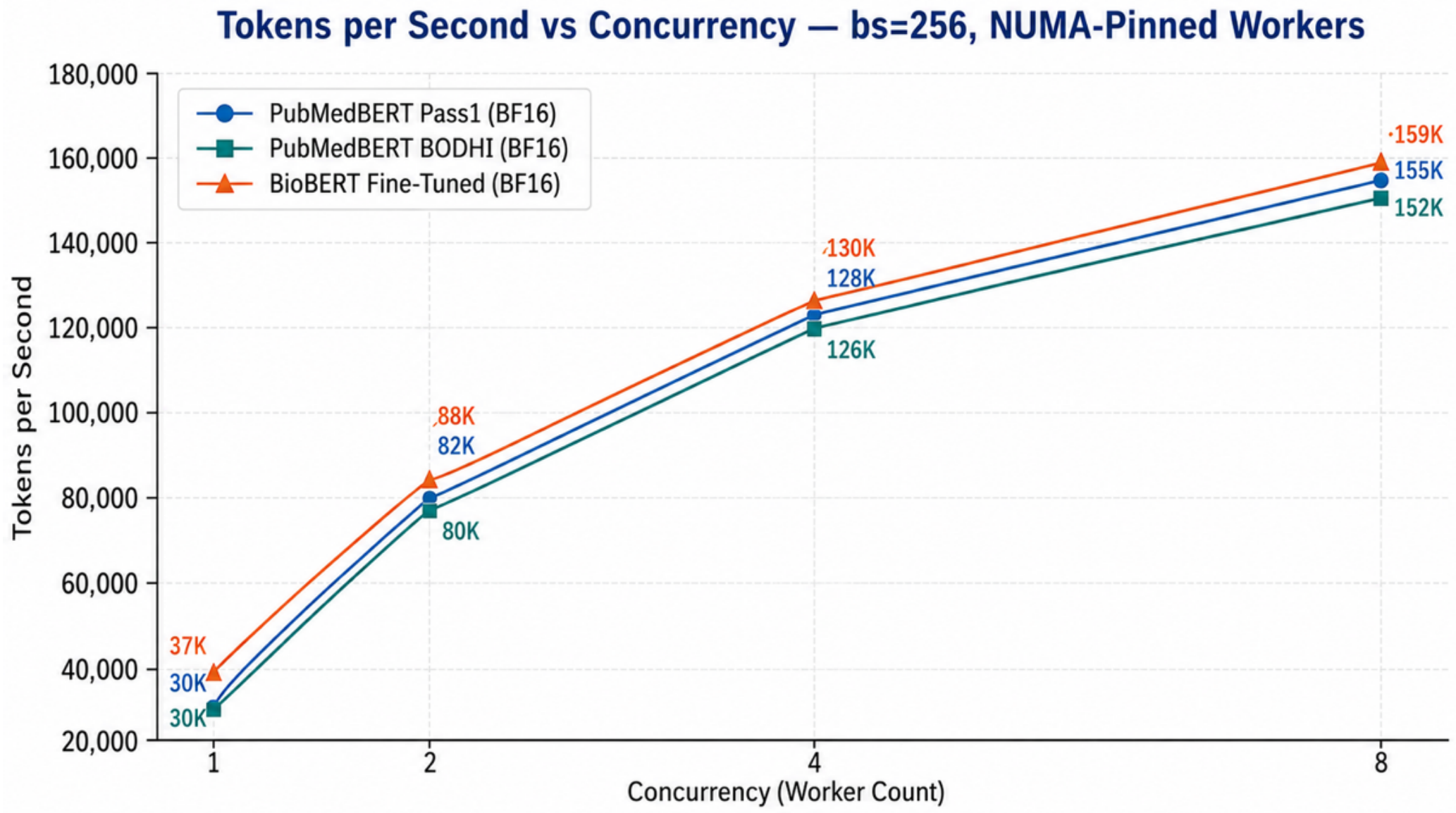

*Figure 14. Batch ingestion. Throughput against worker count, NUMA-pinned, at batch 256. Near-linear to four workers; the eighth worker gives less because memory bandwidth, not compute, becomes the limit. PubMedBERT Pass 1 reaches 11,590 sentences/sec at eight workers.*

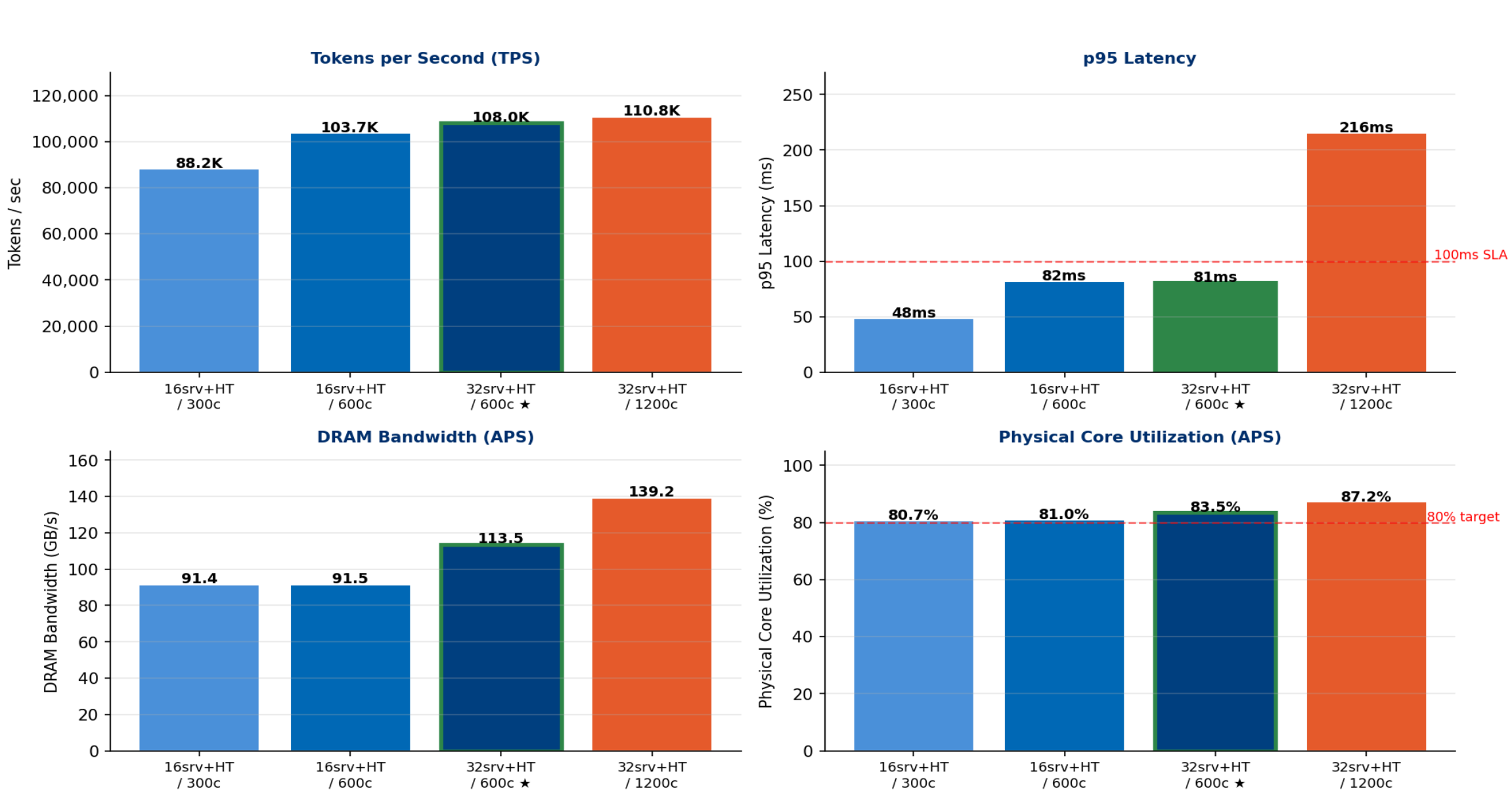

*Figure 15. Online serving, the production sweet spot. Across throughput, p95 latency, DRAM bandwidth, and core utilisation, the 32-server configuration with hyperthreading at 600 clients is the peak-efficiency point: 108k tokens/sec at 80.7 ms p95, 83.5% core utilisation. Pushing to 1,200 clients buys 2.5% more throughput and triples p95.*

The 32srv+HT/600c configuration is the one to ship. It pins 32 OpenVINO servers to two physical plus two hyperthread cores each, which exactly fills 128 logical CPUs. Going wider starves each worker of its hyperthread pair and loses the speedup. The Intel APS profile confirms the machine is well fed without being choked: utilisation rises, IPC peaks at 0.36, and NUMA-remote access stays at 0.6%, so the pinning is doing its job.

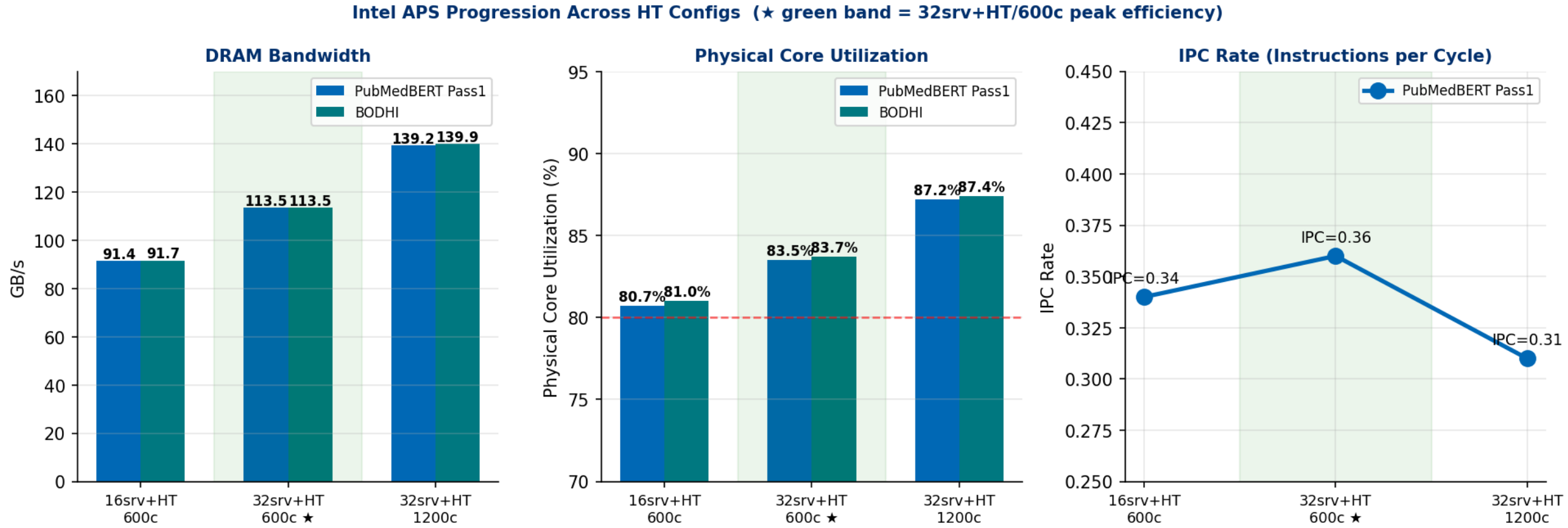

*Figure 16. Hardware counters across three serving configurations. The 32srv+HT/600c point (green band) holds the best balance — higher DRAM bandwidth and utilisation than the 16-server baseline, better IPC than the overloaded 1,200-client run.*

Production sweet-spot scorecard

**PubMedBERT Pass1**

**135,668**

TPS

p50 57 ms · p95 92 ms · cosine 0.9972

*OV INT8 · 32srv+HT · 600c*

**PubMedBERT BODHI**

**135,721**

TPS

p50 58 ms · p95 91 ms · cosine 0.9974

*OV INT8 · 32srv+HT · 600c*

**BioBERT Fine-Tuned**

**182,468**

TPS

p50 67 ms · p95 74 ms · cosine 0.9949

*OV INT8 · 32srv+HT · 600c*

*Figure 17. Production scorecard at the chosen configuration. PubMedBERT Pass 1 and BODHI both serve roughly 135k tokens/sec at ~90 ms p95; the fine-tuned BioBERT reaches 182k. Cosine fidelity stays above 0.994 throughout.*

| Config | Clients | Req/s | TPS | p50 | p95 | DRAM GB/s | Util | IPC |
|---|---|---|---|---|---|---|---|---|
| 8srv / 300c | 300 | 6,579 | 88,227 | 45.5 | 48.3 | 67 | 69.8% | 0.34 |
| 16srv+HT / 600c | 600 | 7,736 | 103,740 | 77.4 | 82.0 | 91.4 | 80.7% | 0.34 |
| 32srv+HT / 600c | 600 | 8,057 | 108,046 | 75.6 | 80.7 | 113.5 | 83.5% | 0.36 |
| 32srv+HT / 1200c | 1200 | 8,265 | 110,840 | 145.2 | 215.5 | 139.2 | 87.2% | 0.31 |

*Table 8. Serving configurations, PubMedBERT Pass 1, batch 256 with a 10 ms dynamic-batching window. The 32srv+HT/600c row is the recommended operating point.*

## 10. How this plugs into the LBM

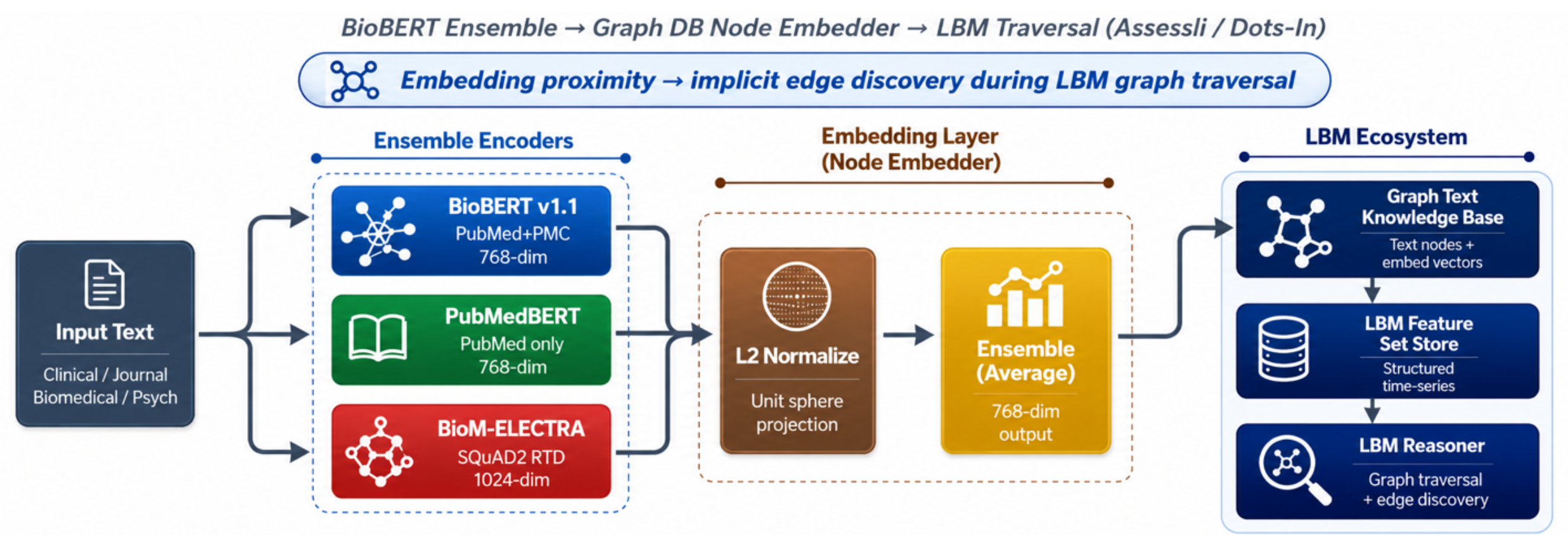

*Figure 18. Where the embedding layer sits. The fine-tuned ensemble embeds every text node in the LBM's graph store; the LBM walks the graph using explicit edges, its own causal priors, and embedding proximity, drawing new edges where proximity suggests a link that does not yet exist.*

The LBM keeps two stores. One is a structured feature store of timestamped measurements — labs, assessment scores, physiological readings — that the model ingests straight into its forward pass. The other is the graph store, where every text node carries the 768-d vector produced by the model in this paper. The LBM does not run nearest-neighbour search over that graph. It reasons over it, walking outward from a seed node along three signals at once: edges a human or a prior pass already drew, the model's own learned sense of which causal chains are plausible, and embedding proximity. When proximity says two unconnected nodes belong together, the model decides whether to draw the edge.

So the embedding quality is the graph quality, directly. With pretrained embeddings the LBM is a non-starter: at 0.76–0.92 cross-domain similarity it would hallucinate an edge on nearly every walk. Pass 1, at 1.63× separation, is usable with real false-edge risk. BODHI, at 2.30×, clears the 1.5× bar where proximity becomes a signal you can lean on. The thresholds we measured — 0.40 for BODHI, 0.50 for Pass 1 — are the dials that set how aggressively the LBM draws edges. Use 0.40 when a false edge is expensive and you want precision; use 0.50 when coverage matters more.

| Use case | Model | Threshold | Platform |
|---|---|---|---|
| Highest STS agreement | PubMedBERT Pass 1 | 0.50 | AMX BF16 |
| Cleanest domain separation | PubMedBERT BODHI | 0.40 | AMX BF16 |
| General biomedical embedding | BioBERT fine-tuned | 0.45 | AMX or c6i |
| Bulk ingestion (>500 s/s) | PubMedBERT Pass 1 | batch 512 | AMX BF16 |
| Low-latency API (<15 ms) | any OV-FP16 model | 0.40 | AMX BF16 |

*Table 9. Pick a model and a threshold by what you are optimising for.*

## 11. What this means, and what it doesn't

The cross-domain failure is not a quirk of biomedical text. It is what masked-token pretraining does to any embedding space asked to bridge vocabularies along a causal chain. Personal finance and mood. Sleep and exam scores. Air quality and breathing. Wearable data and a clinical decision. Anywhere a model has to connect a cause in one domain to an effect in another, it hits the same wall at roughly the same 0.75 cosine. As AI systems move from answering questions to keeping a persistent model of one person's life, this stops being a quality issue and becomes a safety one. A system that infers the wrong causes in someone's life and acts on them is a worse failure than a search box that returns the wrong page.

BODHI generalises. The recipe — positives from edges that exist, hard negatives from edges that do not — works for any domain with a curated ontology. Human Phenotype Ontology, the DSM-5 structure, UMLS, Gene Ontology, and the behavioural ontologies we are building for the LBM are all candidates. Given the size of the Pass 1 to BODHI jump, a second ontology-graph pass looks like a high-leverage move for any embedding model headed for graph reasoning, not just ours.

We are careful about what we are not claiming.

- ELECTRA is the wrong architecture for this. After fine-tuning its gap is +0.066, an order of magnitude short of the BERT models. QA pretraining does not reshape into an embedder. The negative result is the point of including it.
- The benchmark leans biomedical. It covers genetics, biomarkers, physiology, psychology, journals, clinical notes, and finance, but environmental, social, occupational, and wearable-signal domains are thin. Widening it is in progress.
- BODHI today covers biomedical concepts. Extending the absent-edge trick to behavioural and environmental ontologies needs those graphs, some of which exist and some of which we are building.
- The cross-platform comparison uses one no-AMX baseline. Other silicon without dedicated BF16 matrix hardware will land nearer the c6i than the AMX server, but we have not measured AMD, ARM, or Apple parts.
- The 133× figure is single-query latency on AMX. Other regimes give other multipliers; bulk throughput numbers are reported separately and are smaller multiples.

And we are not claiming the LBM is a validated clinical or behavioural intervention — that needs longitudinal user studies we have not run. We are claiming that the embedding substrate underneath it can be made correct, that we measured how correct, and that it runs on hardware you can rent.

## 12. What we are releasing

All of the following ship under a permissive licence:

- The B1–B6 benchmark suite as a Python package with reproducible corpora and one-line entry points.
- The 72,034-pair Pass 1 corpus, labelled by provenance so manual and synthetic pairs are distinguishable.
- The BODHI generator, which turns any ontology graph into (anchor, positive, hard-negative) triplets by walking present and absent edges.
- The MatryoshkaLoss + MNRL training pipeline with the exact hyperparameters used here.
- The OpenVINO conversion scripts and the FP16-versus-INT8 measurement harness.

- The 32srv+HT/600c serving template with start-up scripts, dynamic-batching parameters, and APS hooks.
- Weights for BioBERT fine-tuned, PubMedBERT Pass 1, and PubMedBERT BODHI on Hugging Face, with model cards.

Code at github.com/dotsin/lbm-embedding-bench, weights at huggingface.co/assessli. We want the benchmark extended to new domains and the BODHI trick applied to new ontologies, and we will take pull requests. The LBM's own architecture — its forward pass and its training over the structured feature store — is not in this release and is covered by Indian Patent Application No. 202431040192 (filed, not granted), which concerns the LBM's use in education. The embedding layer is upstream of that, and we are opening it because the failure it exposes is the field's problem to fix, not a moat to defend.

## 13. Conclusion

We started with a model that thinks cortisol and the stock market are 83% alike, and ended with one that keeps the unrelated apart and the genuinely linked together, cleanly enough to reason on. The pretrained encoders fail cross-domain discrimination completely, and they fail because their embedding spaces are anisotropic. A two-pass contrastive fine-tune fixes it: the first pass on 72,034 pairs, the second — BODHI — on the absent edges of a knowledge graph, which is the pass that does the heavy lifting, taking separation to 2.30× and F1 to 93%. And the whole thing runs on commodity Intel CPUs, 133× faster through OpenVINO, with FP16 beating INT8 against every deployment guide's advice.

On populations, getting from correlation to causation is a matter of randomised trials and careful inference. On one person — which is where an LBM lives — it turns out to be, in large part, a matter of embedding geometry. Get the cross-domain pulls and pushes right and the graph traversal can do the rest; get them wrong and no amount of downstream cleverness recovers. This is the layer that gets them right, benchmarked honestly, on hardware that scales to the people the model is meant to serve.

## Acknowledgements

We thank the Intel India AI ISV Partnerships team — Sonali Singh, Satish Kumar Reddy, Maya Vijayan, Srighakollapu Varsha, and Nikhila Haridas — for access to the Xeon 6737P AMX server, infrastructure support, and review of the OpenVINO and VTune profiling work. Soudeep Shaw and Suman Paul of the Assessli engineering team handled systems setup, NUMA pinning, and benchmark runs. We thank dmis-lab, Microsoft Research, and the BioM-ELECTRA authors for the pretrained models, and the Indian Statistical Institute, Kolkata, and NSRCEL at IIM Bangalore for institutional support of the broader LBM programme. The work ran under a confidential NDA between Assessli (ONE OATH Educational Research & Technologies Pvt. Ltd., CIN U80903WB2022PTC256107) and Intel. Compute was provided in-kind by Intel India; no cash grant funded these specific experiments.

## Disclosure

The authors are affiliated with Assessli and Dots-In, product brands of ONE OATH Educational Research & Technologies Pvt. Ltd. (CIN U80903WB2022PTC256107), the developer of the Large Behavioural Model. The benchmark suite, fine-tuning pipelines, and conversion scripts are released open-source as described in Section 12. The LBM's internal architecture is the subject of Indian Patent Application No. 202431040192, covering its application to education. Intel India provided hardware access under NDA and reviewed the disclosure of hardware configuration and performance figures before release; Intel did not direct or edit the scientific content. The authors declare no other competing interests. Contributions: S.B. conceived the LBM architecture, defined the embedding correctness requirement, and designed the BODHI training mechanism; S.G. led fine-tuning, OpenVINO conversion, and deployment; P.M. built the benchmark suite, ran all measurements on both platforms, and led the analysis. All three wrote the paper.